\documentclass[10pt,twocolumn,letterpaper]{article}

\usepackage{cvpr}
\usepackage{times}
\usepackage{epsfig}
\usepackage{graphicx}
\usepackage{amsmath}
\usepackage{amssymb}
\usepackage{enumitem}

\usepackage{bbm}
\usepackage{caption} 
\usepackage{tabularx}  % full wdith tables
\usepackage{booktabs}  % prettier tables
\usepackage[breaklinks=true,bookmarks=false,colorlinks]{hyperref}

\cvprfinalcopy % *** Uncomment this line for the final submission

\def\cvprPaperID{7029} % *** Enter the CVPR Paper ID here

\ifcvprfinal\fi
\begin{document}

%%%%%%%%% TITLE
\title{FBNetV2: Differentiable Neural Architecture Search \\for Spatial and Channel Dimensions} 
\author{
Alvin Wan$^{1}$\thanks{Work done while interning at Facebook.}, Xiaoliang Dai$^{2}$, Peizhao Zhang$^{2}$, Zijian He$^{2}$, Yuandong Tian$^{2}$, Saining Xie$^{2}$, Bichen Wu$^{2}$, \\
Matthew Yu$^{2}$, Tao Xu$^{2}$, Kan Chen$^{2}$, Peter Vajda$^{2}$, Joseph E. Gonzalez$^{1}$\\
$^{1}$UC Berkeley, $^{2}$Facebook Inc.\\
{\tt\small \{alvinwan,jegonzal\}@berkeley.edu}\\
{\tt\small \{xiaoliangdai,stzpz,zijian,yuandong,s9xie,bichen,mattcyu,xutao,kanchen18,vadjap\}@fb.com}
}

\maketitle
%\thispagestyle{empty}

%%%%%%%%% ABSTRACT

% TODO: UPDATE THE NUMBERS AFTER RUNNING L2
\begin{abstract}
Differentiable Neural Architecture Search (DNAS) has demonstrated great success in designing state-of-the-art, efficient neural networks. However, DARTS-based DNAS's search space is small when compared to other search methods', since all candidate network layers must be explicitly instantiated in memory. To address this bottleneck, we propose a memory and computationally efficient DNAS variant: DMaskingNAS.  This algorithm expands the search space by up to $10^{14}\times$ over conventional DNAS, supporting searches over spatial and channel dimensions that are otherwise prohibitively expensive: input resolution and number of filters. We propose a masking mechanism for feature map reuse, so that memory and computational costs stay nearly constant as the search space expands. Furthermore, we employ  effective shape propagation to maximize per-FLOP or per-parameter accuracy. The searched FBNetV2s yield state-of-the-art performance when compared with all previous architectures.
With up to 421$\times$ less search cost, DMaskingNAS finds models with 0.9\% higher accuracy, 15\% fewer FLOPs than MobileNetV3-Small; and with similar accuracy but 20\% fewer FLOPs than Efficient-B0. Furthermore, our FBNetV2 outperforms MobileNetV3 by 2.6\% in accuracy, with equivalent model size. FBNetV2 models are open-sourced at \hyperlink{https://github.com/facebookresearch/mobile-vision}{\color{blue}{https://github.com/facebookresearch/mobile-vision}}.

% \color{blue}{https://github.com/facebookresearch/mobile-vision}

\vspace{-0.2in}

% With similar or higher accuracy, FBNetV2s achieve up to 19\% and 24\% FLOP reduction compared to MobileNetV3 and MnasNet, with more than 421$\times$ less search cost. 
%For example, FBNetV2 achieves 75.7\% top-1 accuracy with only 238M FLOPs.  
%
%For example, with 15\% and 19\% less FLOPs, FBNetV2s achieve 0.9\% accuracy gain and same level of accuracy compared to MobileNetV3.
%
%For example, with 15\% and 11\% less FLOPs, FBNetV2s achieve 0.9\% and 4.1\% accuracy improvements compared to MobileNetV3 and MnasNet, respectively, with more than 421$\times$ less search cost.
%

\end{abstract}

%%%%%%%%% BODY TEXT
\section{Introduction}
Deep neural networks have led to significant progress in many research areas and applications, such as computer vision and autonomous driving. Despite this, designing an efficient network for resource-constrained settings remains a challenging problem. Initial directions involved compressing existing networks~\cite{deepcompression} or building small networks~\cite{shufflenetv2, mobilenetv2}. However, the design space can easily contain more than $10^{18}$ candidate architectures~\cite{fbnet, single_path}, making manual design choices sub-optimal and difficult to scale. In lieu of manual tuning, recent work uses neural architecture search (NAS) to design networks automatically.

\begin{figure}[t]
    \centering
    \includegraphics[width=0.48\textwidth]{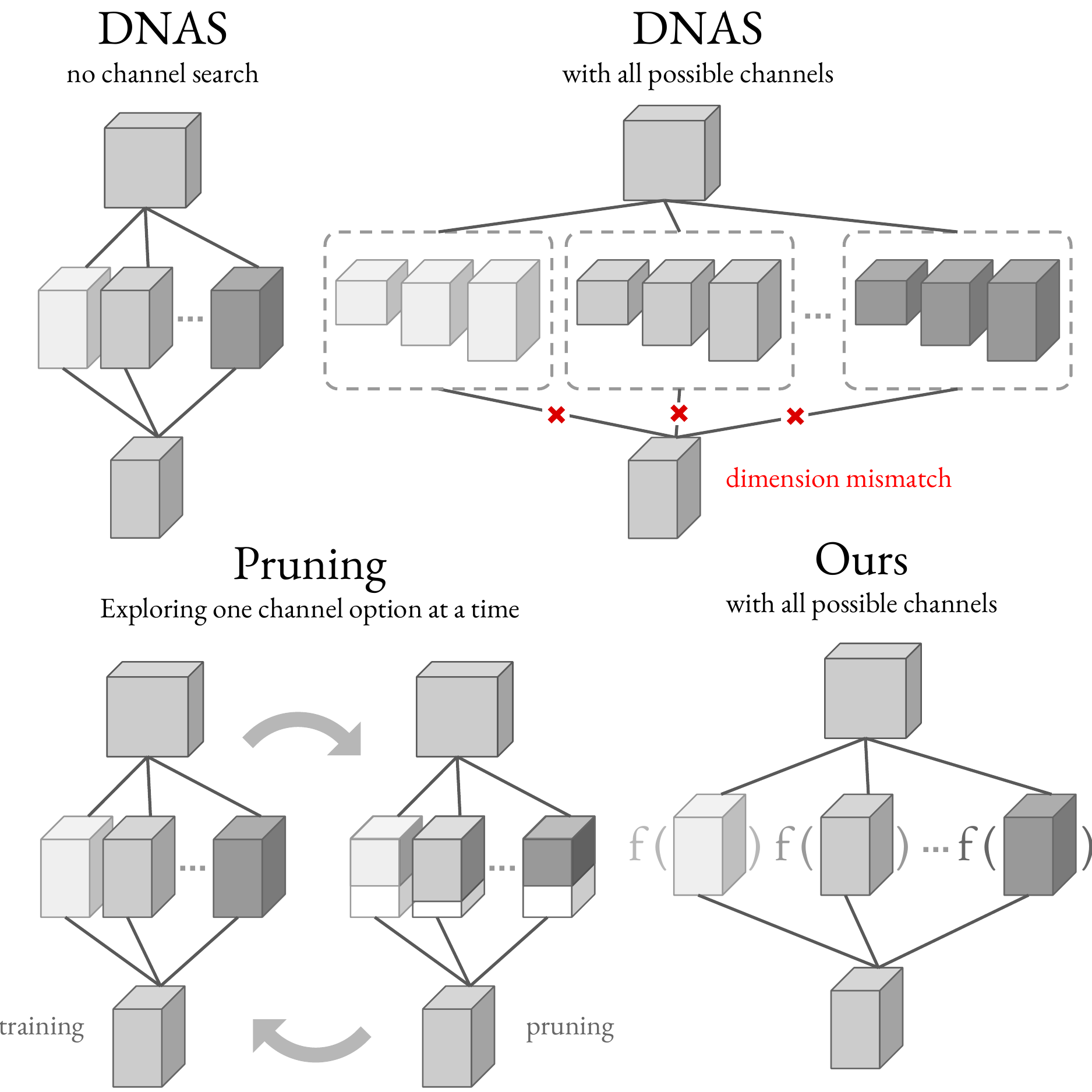}
    \caption{\textbf{DNAS}: Adding all possible numbers of filters to DNAS (top-right) increases computational and memory costs drastically, exacerbating DNAS's memory bottleneck on search space size. \textbf{Pruning}: Channel pruning (bottom-left) is limited to training one architecture at a time. \textbf{Ours}: With our weight-sharing approximation, DNAS can explore all possible number of filters simultaneously with negligible memory and computation overhead. See Fig.~\ref{fig:method} for details.}
    \label{fig:main}
\end{figure}

Previous NAS methods utilize reinforcement learning (RL) techniques or evolutionary algorithms (EAs). However, both methods are computationally expensive and consume thousands of GPU hours~\cite{nasnet, mnasnet}. As a result, recent NAS literature~\cite{fbnet, darts, enas} focuses on differentiable neural architecture search (DNAS); DNAS searches over a supergraph that encompasses all candidate architectures, selecting a single path as the final neural network. Unlike conventional NAS, DNAS can search large combinatorial spaces in the time it takes to train a single model~\cite{darts, snas, fbnet, single_path}. One class of DNAS methods, based on DARTS~\cite{darts}, suffer from two significant limitations~\cite{nassurvey}:

\begin{itemize}
    \item \textbf{Memory costs bound the search space.} Short of paging in and out tensors, the supergraph and feature maps must reside in GPU memory for training, which limits the search space.
    \item \textbf{Cost grows linearly with the number of options per layer.} This means that each new search dimension introduces combinatorially more options and combinatorial memory and computational costs.
\end{itemize}

\begin{figure*}
    \centering
    \includegraphics[width=\textwidth]{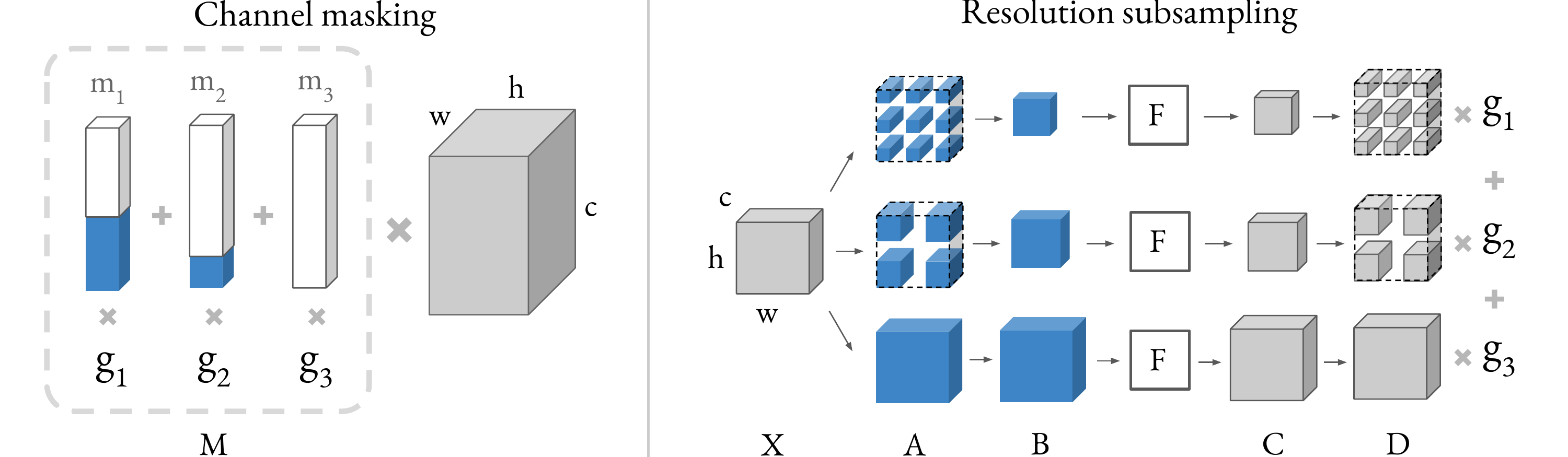}
    \caption{\textbf{Channel Masking} for channel search: A column vector mask $M \in \mathbb{R}^c$ is the weighted sum of several masks $m_i \in \mathbb{R}^c$, with Gumbel Softmax weights $g_i$. Each $m_i$ has ones (white) in the first $k$ entries and zeros (blue) in the next $c-k$ entries, for some $k \in \mathbb{Z}$. Multiplication with this mask speeds up channel search, using a weight-sharing approximation described in Fig.~\ref{fig:channel_challenges}. \textbf{Resolution Subsampling} for input resolution: $X$ is an intermediate output feature map for the network. $A$ is subsampled from $X$ using nearest neighbors. Values at the blue pixels in column $A$ are assembled to create the smaller feature map in $B$. Next, run the operation $F$. Finally, each value in $C$ is placed back into a larger feature map in $D$. Note we put values back ($D$) into pixels where we pulled values from ($A$). This process is motivated in Fig.~\ref{fig:spatial_challenges}.}
    \label{fig:method}
\end{figure*}

The other class of DNAS methods, not based on DARTS, suffer from similar issues: For example, ProxylessNAS tackles the memory constraint by training only one path in the supergraph each iteration. However, this means ProxylessNAS would take a prohibitively long time to converge on an order-of-magnitude larger search space. These memory and computation issues, for all DNAS methods, prevent us from expanding the search space to explore larger spaces of configurations. Noting that feature maps typically dominate memory cost~\cite{nvda_guide}, we propose a formulation of DNAS (Fig.~\ref{fig:main}) called DMaskingNAS (Fig.~\ref{fig:method}) that increases the search space size by orders of magnitude. To accomplish this, we represent multiple channel and input resolution options in the supergraph with masks, which carry negligible memory and computational costs. Furthermore, we reuse feature maps for all options in the supergraph, which enables nearly constant memory cost with increasing search space sizes. These optimizations yield the following three contributions:

% TODO(alvin): add limiting case somewhere
\begin{itemize}
    \item \textbf{A memory and computationally efficient DNAS} that optimizes both macro- (resolution, channels) and micro- (building blocks) architectures jointly in a $10^{14}\times$ larger search space using differentiable search. To the best of our knowledge, we are the first to tackle this problem using a differentiable search framework supergraph, with substantially less computational cost and roughly constant memory cost.
    \item \textbf{A masking mechanism and effective shape propagation} for feature map reuse. This is applied to both the spatial and channel dimensions in DNAS.
    \item \textbf{State-of-the-art results} on ImageNet classification. With only 27 hours on 8 GPUs, our searched compact models lead to substantial per-parameter, per-FLOP accuracy improvements. The searched models outperform all previous state-of-the-art neural networks, both manually and automatically designed, small and large.
\end{itemize}

\begin{table}[ht] \small
\centering
\caption{The number of DMaskingNAS design choices eclipses that of previous search spaces: number of channels $c$, kernel size $k$, number of layers $l$, bottleneck type $b$, input resolution $r$, and expansion rate $e$.}

\begin{tabular*}{0.48\textwidth}{l@{\extracolsep{\fill}}llllll}
\toprule
NAS algorithm                       & c & k & l & b & r & e \\
\midrule
MnasNet~\cite{mnasnet}              &\checkmark
  &  &\checkmark  &\checkmark  &  &\checkmark    \\
ProxylessNAS~\cite{proxylessnas}    &   &\checkmark  &\checkmark  &\checkmark  &   &\checkmark   \\
Single-Path NAS~\cite{single_path}  &   &\checkmark  &\checkmark  &   &   &\checkmark   \\
ChamNet~\cite{chamnet}              &\checkmark  &   &\checkmark  &   &\checkmark  &\checkmark    \\
FBNet~\cite{fbnet}                  &   &\checkmark  &\checkmark  &\checkmark  &   &\checkmark  \\
DMaskingNAS                         &\checkmark  &\checkmark  &\checkmark  &\checkmark  &\checkmark  &\checkmark  \\
\bottomrule
\end{tabular*}
\label{tab:searchspace}
\vspace{-0.2in}
\end{table}

\section{Related Work}

Hand-crafted, efficient neural networks see two predominant approaches: (1) compressing existing architectures and (2) designing compact architectures from scratch.

\textbf{Network compression} includes both architectural and non-architectural modifications. One non-architectural approach is low-bit quantization, where weights and activations alike may be represented with fewer bits.  For example, Wang et al.~\cite{haq} propose hardware-aware automated quantization, which achieves a $1.4$-$1.95\times$ latency reduction on MobileNet~\cite{mobilenet}. These techniques are orthogonal to and can be combined with the methods in this paper. Alternatively, architectural modifications include network pruning~\cite{netprune, structured_sparsity, energy_aware}, where various heuristics govern layer-wise or channel-wise pruning. For example, Han et al.~\cite{netprune} show that magnitude-based pruning can reduce parameter count by orders of magnitude without accuracy loss, and NetAdapt~\cite{netadapt} utilizes a filter pruning algorithm that achieves a 1.2$\times$ speedup for MobileNetV2. However, with heuristics-based simplifications, pruning methods train potential architectures separately, one after another -- in some cases, pruning methods consider only one architecture \cite{liu2017slimming, he2018sfp}.

\textbf{Compact architecture design} aims to directly construct efficient networks, rather than trim an expensive one~\cite{squeezenet, squeezedet}.  For example, MobileNet~\cite{mobilenet} and MobileNetV2~\cite{mobilenetv2} achieve substantial efficiency improvements by exploiting a depth-wise convolution and an inverted residual block, respectively.  ShuffleNetV2~\cite{shufflenetv2} shrinks the model size utilizing low-cost group convolutions. Tan et al. propose a compound scaling method, obtaining a family of architectures that achieve state-of-the-art accuracy with an order of magnitude fewer parameters than previous convolutional networks~\cite{efficientnet}. However, these models rely on finely-tuned, manual decisions that are bested by automatic design.

\textbf{Neural architecture search} automates the design of state-of-the-art neural networks. Zoph et al. first proposed using RL for automated neural network design in~\cite{NASRL}. This and other early NAS approaches are based on RL~\cite{NASRL, mnasnet} and EA~\cite{evolution}. However, both approaches consume substantial computational resources.

Later works utilize various techniques to reduce the computational cost of search. One such technique formulates the architecture search problem as a path-finding process in a supergraph ~\cite{fbnet, darts, one-shot, single_path}. Among them, gradient-based NAS has emerged as a promising tool. Wu et al. show that gradient-based, differentiable NAS yields state-of-the-art compact architectures with $421\times$ less search cost than RL-based approaches. %However, supergraph-based methods suffer from severely limited search space sizes, due to memory constraints.
Another direction is to exploit a performance predictor to guide the search process~\cite{chamnet, progressive}. Such approaches explore the search space by trimming progressively and lead to significant reductions in search cost. 

Stamoulis et al.~\cite{singlepathnas} introduce weight-sharing to further reduce the computational cost of search. However, kernel weight-sharing doesn't address the primary drawback of DARTS, namely a memory bottleneck yielding small search space size: Say a ``mixed kernel" contains weights shared between a $3\times3$ and $5\times5$. Since it is impossible to extract a $3\times3$ convolution's outputs from a $5\times5$'s (and vice versa), this mixed kernel still convolves $2\times$ and still stores 2 feature maps for backpropagation. Thus, 2 kernel-weight-sharing convolutions induce memory and computational costs of 2 vanilla convolutions.

% Note: performance predictor approach typically doesn't rely on a supergraph
% also, single_path actually proposes to mask the weights thus to search for kernels efficiently (in a single path).  Do we need to explain more about their work?

\textbf{Searching along spatial and channel dimensions} has been studied both with and without NAS. Liu et al~\cite{autodeeplab} develop a NAS variant that searches over varying strides for semantic segmentation. However, this method suffers from increasing memory cost as the number of possible input resolutions grows. As described above, network pruning suffers from inefficient and sequential exploration of architectures, one-by-one. Yu et al~\cite{slimmable} amend this partially by creating a batchnorm invariant to the number input channels; after training the ``supergraph" they see competitive accuracy without further training, for each possible subset of channels. Yu et al~\cite{autoslim} expand on these slimmable networks by introducing a test-time greedy channel selection procedure. However, these methods are orthogonal to and can be combined with DMaskingNAS, as we train the sampled architecture from scratch. To address these concerns, our algorithm jointly optimizes over multiple input resolutions and channel options simultaneously, increasing memory cost only negligibly as the number of options grows. This allows DMaskingNAS to support orders of magnitude more possible architectures, under existing memory constraints.

\section{Method}

We propose DMaskingNAS to search over spatial and channel dimensions, summarized in Fig.~\ref{fig:method}. 
The search space would be computationally prohibitive and ill-formed without the optimizations described below; our approach makes it possible to search this expanded search space (Table~\ref{tab:searchspace}) over channels and input resolutions.

\subsection{Channel Search}\label{sec:channel-search}

\begin{figure}
    \centering
    \includegraphics[width=0.48\textwidth]{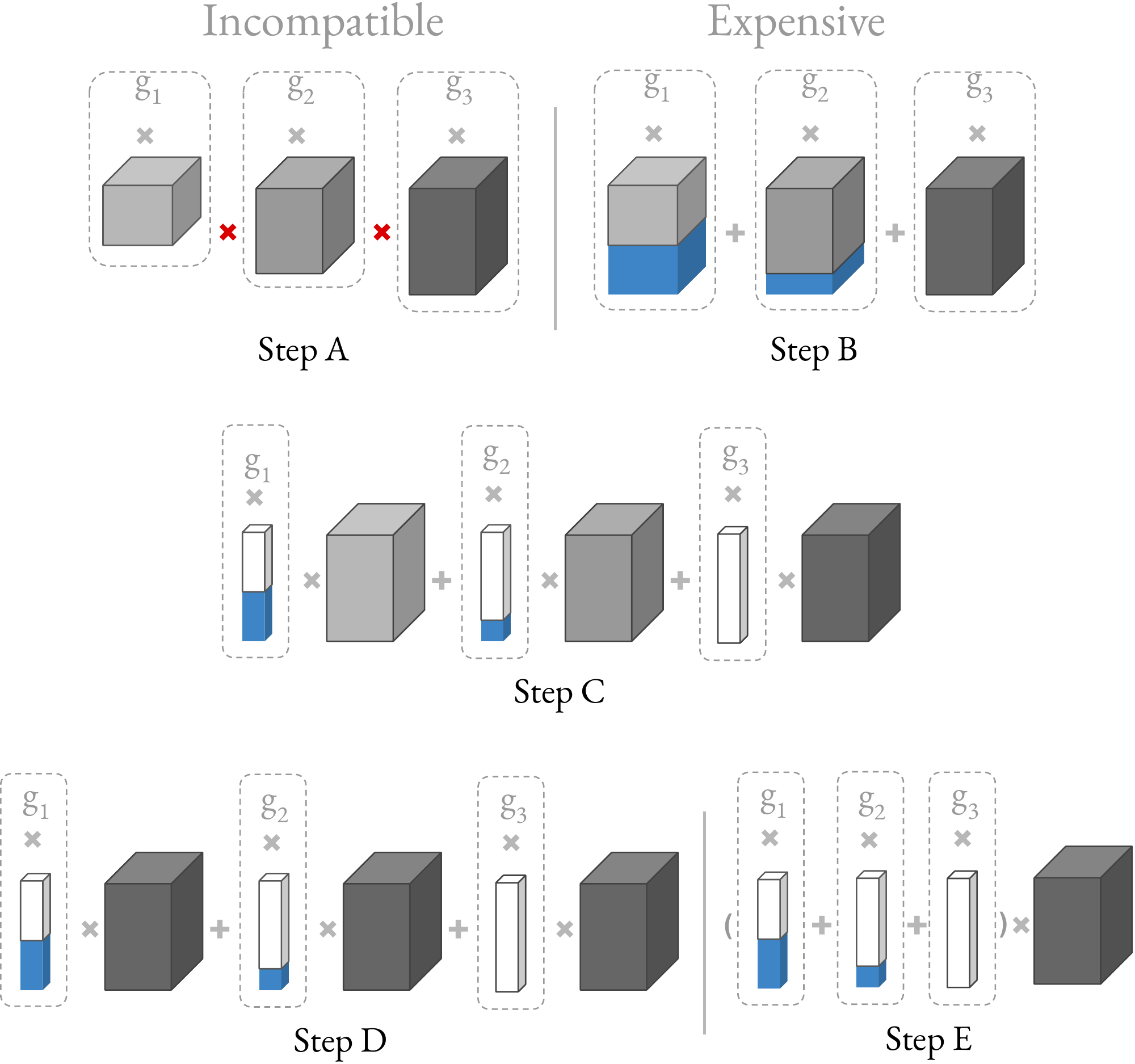}
    \caption{\textbf{Channel Search Challenges}: \textbf{Step A}: Consider 3 convolutions with varying numbers of filters. Each output (gray) will have varying numbers of channels. Thus, the outputs cannot be naively summed. \textbf{Step B}: Zero-padding (blue) outputs allows them to be summed. However, both FLOP and memory cost increases sub-linearly with the number of channel options. \textbf{Step C}: This is equivalent to running three convolutions with equal numbers of filters, multiplied by masks of zeros (blue) and ones (white). \textbf{Step D}: We approximate using weight sharing -- all three convolutions are represented by one convolution. \textbf{Step E}: This is equivalent to summing the masks first, before multiplying by the output. Now, FLOP and memory cost are effectively constant w.r.t. the number of channel options. % This masking is re-illustrated in Fig. \ref{fig:method}, Channel Masking.
    }
    \label{fig:channel_challenges}
\end{figure}

To support searches over varying numbers of channels, previous DNAS methods simply instantiate a block for every channel option in the supergraph. For a convolution with $k$ filters, this could mean up to $k(k+1)/2\sim O(k^2)$ convolutions. Previous channel pruning methods ~\cite{autoslim} suffer from a similar drawback: each option must be trained separately, finding the ``optimal" channel count in one shot or iteratively.
Furthermore, even without saturating the maximum number of possibilities, there are two problems, the first of which makes this search impossible:

\begin{enumerate}
    \item \textbf{Incompatible dimensions}: DNAS is divided into several ``cells". In each cell, we consider a number of different block options; the outputs of all options are combined in a weighted sum. This means that all block outputs must align dimensions. If each block adopts convolutions with different number of filters, each output will have a different number of channels. As a result, DNAS could not perform a weighted sum.
    \item \textbf{Slower training, increased memory cost}: Even with a workaround, with this naïve instantiation, each convolution with a different channel option must be run separately, resulting in a $O(k)$ increase in FLOP cost. Furthermore, each output feature map must be stored separately in memory.%, at minimum doubling training time. 
     %, at minimum doubling feature map memory cost for this block.
\end{enumerate}

To address the aforementioned issues, we handle the incompatibility (Fig.~\ref{fig:channel_challenges}, Step A): consider a block $b$ with varying numbers of filters, where $b_i$ denotes this block with $i$ filters. The maximum number of filters is $k$. The outputs of all blocks are then zero-padded to have $k$ channels (Fig.~\ref{fig:channel_challenges}, Step B). Given input $x$, the Gumbel Softmax output is thus the following, with Gumbel weights $g_i$:

\begin{equation}\label{eqn:channel-search-1}
    y = \sum_{i=1}^k g_i \textsc{Pad}(b_i(x), k)
\end{equation}
Note that this is equivalent to increasing the number of filters for all convolutions to $k$, and masking out the extra channels (Fig.~\ref{fig:channel_challenges}, Step C). $\mathbbm{1}_{i} \in \mathbb{R}^k$ is a column vector with $i$ leading 1s and $k-i$ trailing zeros. Note that the search method is invariant to the ordering of 1s and 0s. % TODO(alvin) ``proof" in supp
Since all blocks $b_i$ have the same number of filters, we can approximate by sharing weights, so that $b_i = b$ (Fig.~\ref{fig:channel_challenges}, Step D).

\begin{equation}\label{eqn:channel-search-2}
    y = \sum_{i=1}^k g_i (b(x) \circ \mathbbm{1}_{i})
\end{equation}

Finally, with this approximation, we can handle the computational complexity of the naïve channel search approach: this is equivalent to computing the aggregate mask and running the block $b$ only once (Fig.~\ref{fig:channel_challenges}, Step E).

\begin{equation}\label{eqn:channel-search-3}
    y = b(x) \circ \underbrace{\sum_{i=1}^k g_i \mathbbm{1}_{i}}_{M}
\end{equation}

This approximation only requires one forward pass and one feature map, inducing no additional FLOP or memory costs other than the negligible $M$ term in Eq.~\ref{eqn:channel-search-3} (Fig.~\ref{fig:method}, Channel Masking). Furthermore, the approximation falls short of equivalence only because weights are shared, which is shown to reduce train time and boost accuracy in DNAS ~\cite{singlepathnas}. This allows us to search the number of output channels for any block, including related architectural decisions such as the expansion rate in an inverted residual block.

\subsection{Input Resolution Search}

\begin{figure}
    \centering
    \includegraphics[width=0.48\textwidth]{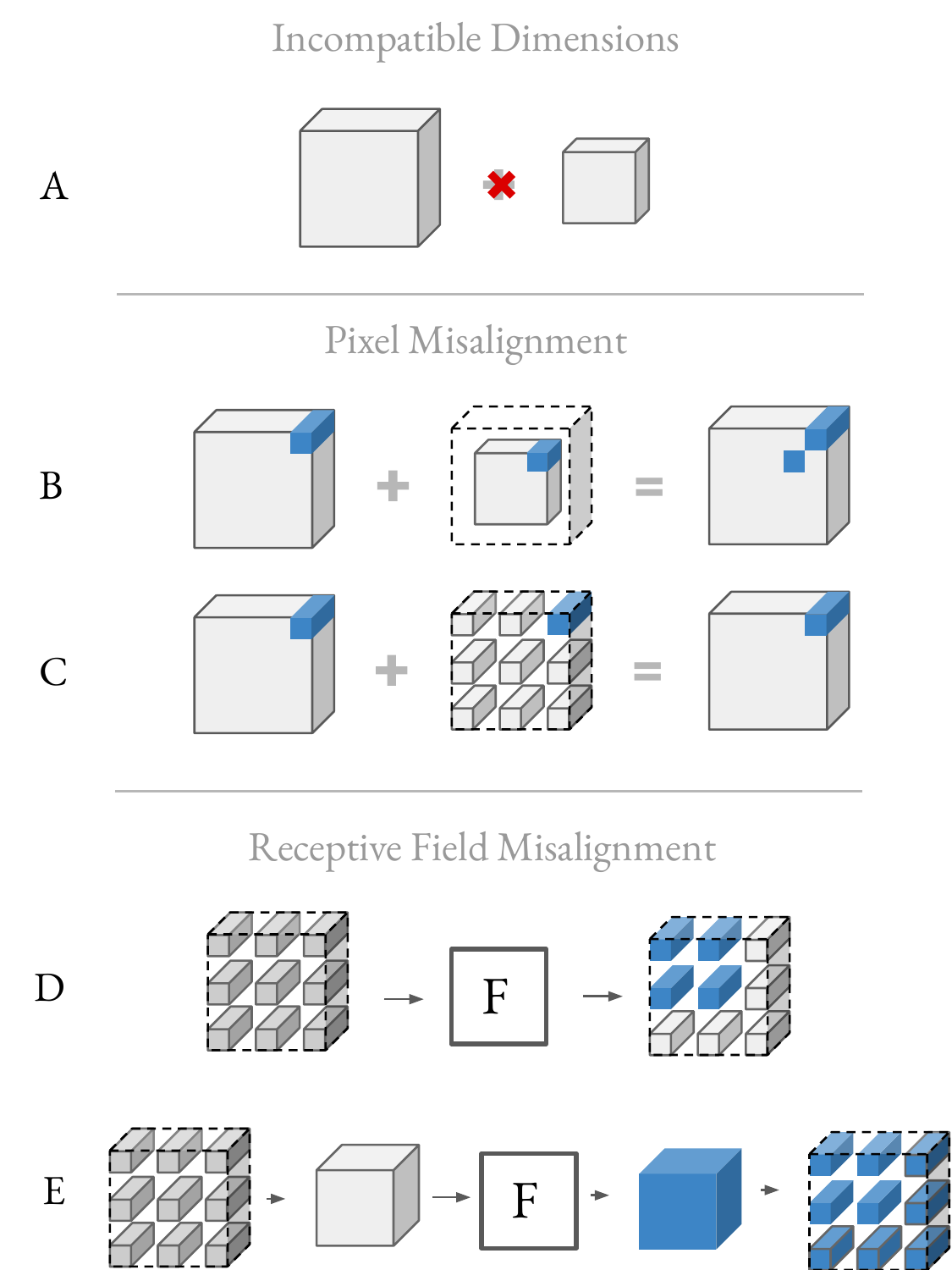}
    \caption{\textbf{Spatial Search Challenges}: \textbf{A}: Tensors with different spatial dimensions cannot be summed due to incompatible dimensions. \textbf{B}: Zero-padding along the periphery of the smaller feature map makes summing possible. However, the top-right pixels (blue) are not aligned correctly. \textbf{C}: Interspersing zero-padding spatially results in a sum that aligns pixels correctly. Note the top-right pixels of both feature maps are correctly overlapping in the sum. \textbf{D}: Say $F$ is a convolution with $3\times3$ kernels. Convolving naïvely with the feature map, containing a subset (gray), results in reduced receptive field ($2\times2$, blue) for the subset. \textbf{E}: To preserve receptive field for all searched input resolutions, the input must be subsampled before convolving. Note the receptive field (blue) is still $3 \times 3$. Furthermore, note we can achieve the same effect, without the need to construct a smaller tensor, with appropriately-strided dilated convolutions; we subsample to avoid modifying the operation $F$.}
    \label{fig:spatial_challenges}
\end{figure}

\begin{figure}
    \centering
    \includegraphics[width=0.48\textwidth]{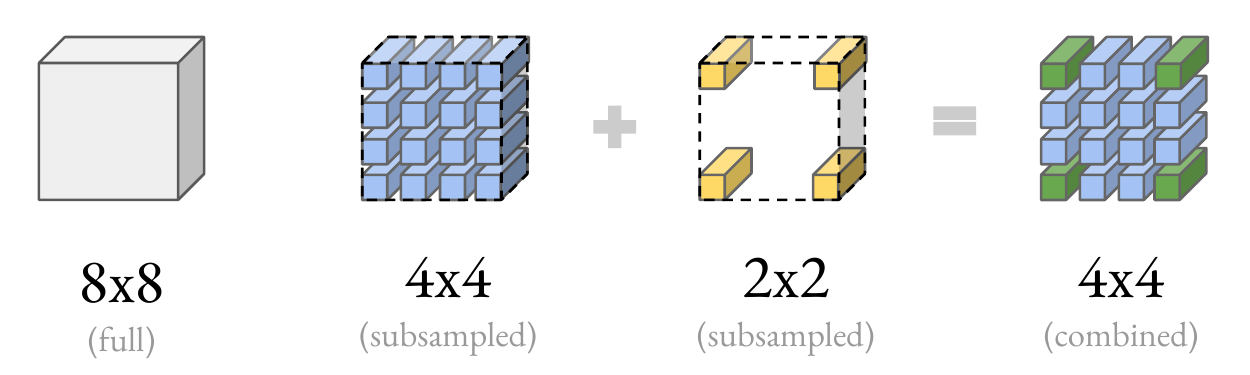}
    \caption{\textbf{Minimizing Pixel ``Contamination"}: On the far left, we have the original $8\times8$ feature map. The blue $4\times4$ is a feature map subsampled with nearest neighbors and zero-padded uniformly. The yellow $2\times2$ is also subsampled and zero-padded. Summing the $2\times2$ with the $4\times4$ yields the combined feature map to the far right. Only the green pixels in the corners hold values from both feature map sizes; these green values are ``contaminated" by the lower resolution feature maps.}
    \label{fig:spatial_challenges_conflicts}
\end{figure}

For spatial dimensions, we search over input resolutions. As with channels, previous DNAS methods would simply instantiate each block with every input resolution. This naïve method's downfalls are twofold: increased memory cost and incompatible dimensions. As before, we address both issues directly by zero-padding the result. However, there are two caveats:

\begin{enumerate}
    \item \textbf{Pixel misalignment}: means padding cannot occur naïvely as before. It would not make sense to zero-pad the periphery of the image, since the sum in Eq.~\ref{eqn:channel-search-1} would result in misaligned pixels (Fig.~\ref{fig:spatial_challenges}, B). To handle pixel misalignment, we zero-pad such that zeros are interspersed spatially (Fig.~\ref{fig:spatial_challenges}, C). This zero-padding pattern is uniform; except for the zeros, this is a nearest neighbors upsampling. For example, a $2\times$ increase in size would involve zero-padding every other row and column. Zero-padding instead of upsampling minimizes ``pixel contamination" across input resolutions (Fig.~\ref{fig:spatial_challenges_conflicts}).
    \item \textbf{Receptive field misalignment}: Since subsets of the feature map correspond to different resolutions, naïvely convolving over the full feature map would result in a reduced receptive field (Fig.~\ref{fig:spatial_challenges}, D). To handle receptive field misalignment, we convolve over subsampled input instead. (Fig.~\ref{fig:spatial_challenges}, E). Using Gumbel Softmax, we arrive at ``resolution subsampling" in Fig.~\ref{fig:method}.
\end{enumerate}

NASNet~\cite{nasnet} introduces a similar notion of combining hidden states. These combinations are also used to efficiently explore a combinatorially large search space but are used to determine -- instead of input resolution or channels -- the number of times to repeat a searched cell. With the above insights, the input resolution search thus incurs constant memory cost, regardless of the number of input resolutions. On the other hand, computational cost increases sub-linearly as the number of resolutions grows.

\subsection{Effective Shape Propagation}

Note this calculation for effective shape is only used during training. In our formulation of the weighted sum Eq.~\ref{eqn:channel-search-1}, the output $y$ retains the maximum number of channels. However, there exists a non-integral number of \textit{effective} channels: say a 16-channel output has Gumbel weight $g_i = 0.8$ and a 12-channel output has weight $g_i = 0.2$. This means the effective number of channels is $0.8 * 16 + 0.2 * 12 = 15.2$. These effective channels are necessary for both FLOP and parameter computation, as assigning higher weight to more channels should incur a larger cost penalty. This effective shape is how we realize effective resource costs introduced in previous works ~\cite{fbnet, snas}: First, define the gumbel softmax weights as

\begin{equation}\label{eqn:gumbel}
g^{l}_{i} = \frac{\text{exp}[(\alpha^{l}_{i} + \epsilon^{l}_{i}) / \tau]}
   {\Sigma_{i} \text{exp}[(\alpha^{l}_{i} + \epsilon^{l}_{i}) / \tau]}\\
\end{equation}

with sampling parameter $\alpha$, Gumbel noise $\epsilon$, temperature $\tau$. For a convolution with Gumbel Softmax in the $l^{th}$ layer, we define its effective output shape $\bar{S}^{l}_{out}$ in Eq.~\ref{eqn:effective_out} using effective output channel ($\bar{C}^{l}_{out}$, Eq.~\ref{eqn:effective_channel}), and effective height, width ($\bar{h}^{l}_{out}, \bar{w}^{l}_{out}$, Eq.~\ref{eqn:effective_spatial}).
\begin{equation}
\label{eqn:effective_channel}
     \bar{C}^{l}_{out} = \Sigma_{i} g^{l}_{i}\cdot C^{l}_{i, out} \\
\end{equation}
\begin{equation}\label{eqn:effective_spatial}
% \begin{split}
     \bar{h}^{l}_{out} = \Sigma_{i} g^{l}_{i}\cdot \bar{h}^{l}_{in}, \bar{w}^{l}_{out} = \Sigma_{i} g^{l}_{i}\cdot \bar{w}^{l}_{in}
% \end{split}
\end{equation}
\begin{equation}\label{eqn:effective_out}
    \bar{S}^{l}_{out} = (n, \bar{C}^{l}_{out}, \bar{h}^{l}_{out}, \bar{w}^{l}_{out})
\end{equation}
with batch size $n$, effective input width $\bar{w}_{in}$ and height $\bar{h}_{in}$.

For a convolution layer without a Gumbel Softmax, effective output shape simplifies to Eq.~\ref{eqn:noneffective_channel}, where effective channel count is equal to actual channel count. For a depth-wise convolution, effective output shape simplifies to Eq.~\ref{eqn:depthwise_conv_cost}, where effective channel count is simply propagated.
\begin{equation}
\label{eqn:noneffective_channel}
\begin{split}
    & \bar{C}^{l}_{out} =  C^{l}_{out} \\
\end{split}
\end{equation}
\begin{equation}
\label{eqn:depthwise_conv_cost}
\begin{split}
    & \bar{C}^{l}_{out} = \bar{C}^{l}_{in} \\
\end{split}
\end{equation}
with actual output channel count $C_{out}$, effective input channel count $\bar{C}_{in}$. Then, we define the cost function for the $l^{th}$ layer as follow:

\begin{equation}
\text{cost}^{l} = 
    \left\{ 
    \begin{aligned}
    & k^{2}\cdot \bar{h}^{l}_{out} \cdot \bar{w}^{l}_{out} \cdot \bar{C}^{l}_{in}\cdot \bar{C}^{l}_{out}\ /\ \gamma & \text{if \texttt{FLOP}}\\
    & k^{2} \cdot \bar{C}^{l}_{in}\cdot \bar{C}^{l}_{out}\ /\ \gamma & \text{if \texttt{param}}
    \end{aligned}
    \right. \\
\end{equation}
with $\gamma$ convolution groups. The effective input channels for the $(l+1)^{th}$ layer are $\bar{C}^{l+1}_{in} = \bar{C}^{l}_{out}$. The total training loss consists of (1) cross-entropy loss and (2) total cost, which is the sum of cost from all layers: 
$\text{cost}_{total} = \Sigma_{l} \text{cost}^{l}$.

In the forward pass, for all convolutions, we calculate and return both the output tensor and effective output shape. Additionally, $\tau$ in the Gumbel Softmax Eq.~\ref{eqn:gumbel} decreases throughout training,~\cite{gumbel}, forcing $g^{l}$ to approach a one-hot distribution. $\text{argmax}_i g^{l}_i$ would thus select a path of blocks in the supergraph; a single channel and expansion rate option for each block; and a single input resolution for the entire network. This final architecture is then trained. Note this final model does not employ masking or require effective shapes. %Then, in the backward pass, we train both the Gumbel Softmax parameters and network weights based on the gradients of the overall loss.

\section{Experiments}

We use DMaskingNAS to search for convolutional network architectures under different objectives.  We compare our search space, performance of searched models, and search cost to previously state-of-the-art networks. Detailed numerical results are listed in Table~\ref{tab:imagenet}.

\subsection{Experimental Setup}
We implement DMaskingNAS using PyTorch on 8 Tesla V100 GPUs with 16GB memory. We use DMaskingNAS to search for convolutional neural networks on the ImageNet (ILSVRC 2012) classification dataset~\cite{imagenet}, a widely-used NAS evaluation benchmark. We use the same training settings as reported in~\cite{fbnet}: we randomly select 10\% of classes from the original 1000 classes and train the supergraph for 90 epochs. In each epoch, we train the network weights with 80\% of training samples using SGD. We then train the Gumbel Softmax sampling parameter $\alpha$ with the remaining 20\% using Adam~\cite{kingma2014adam}. We set initial temperature $\tau$ to 5.0 and exponentially anneal by $e^{-0.045}\approx 0.956$ every epoch.

\subsection{Search Space}
Previous cell-level searches produced fragmented, complicated, and latency-unfriendly blocks. Thus, we adopt a layer-wise search space for known, latency-friendly blocks. 

\begin{figure}
    \centering
    \includegraphics[width=0.46\textwidth]{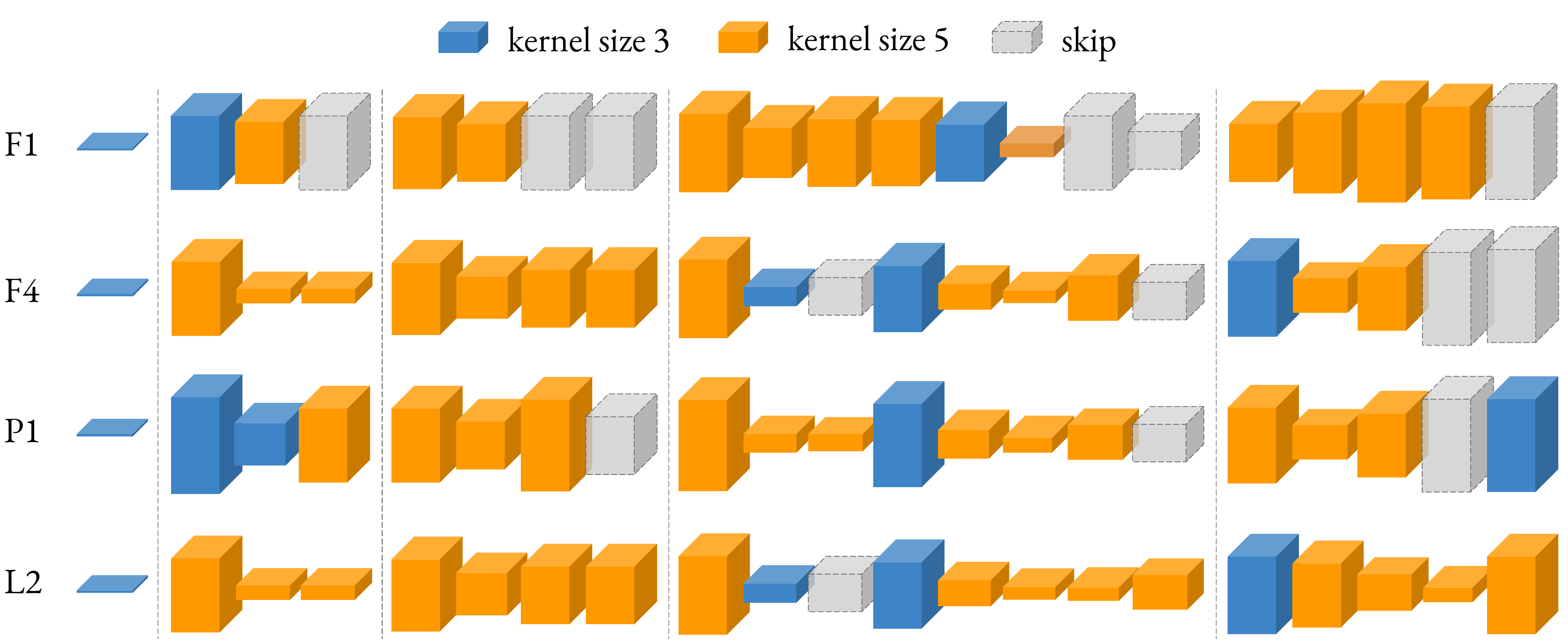}
    \caption{Searched FBNetV2 architectures, with colors denoting different kernel sizes and heights denoting different expansion rates. The heights are drawn to scale.}
    \label{fig:architectures}
\end{figure}

\begin{table}[h]
\footnotesize
\centering
\caption{\textbf{Macro-architecture} for our largest search space, describing block type $b$, block expansion rate $e$, number of filters $f$, number of blocks $n$, stride of first block $s$. ``TBS'' means layer type needs to be searched. Tuples of three values represent the lowest value, highest, and steps between options (low, high, steps). The maximum input resolution for FBNetV2-P models is 288, for FBNetV2-F is 224, and for FBNetV2-L is 256. See supplementary material for all search spaces. % TODO(alvin): add all search spaces to supplementary materials
% TODO(alvin): fix search space dimensions below
}

% \begin{tabular*}{0.48\textwidth}{l @{\extracolsep{\fill}} lllll}
% \toprule
% Max. Input              & b           & e         & f          & n        & s \\ \midrule
% $224^2 \times 3$        & 3x3         & 1         & (8, 16, 8) & 1        & 2           \\
% $112^2 \times 16$       & TBS         & 1         & (8, 16, 8) & 2        & 1      \\
% $112^2 \times 16$       & TBS         & (1, 3)    & (8, 24, 8) & 2        & 1      \\
% $56^2 \times 24$        & TBS         & (1, 3)    & (8, 24, 8) & 2        & 2           \\
% $56^2 \times 24$        & TBS         & (1, 3)    & (8, 64, 8) & 2        & 1      \\
% $28^2 \times 64$        & TBS         & (1, 4)    & (8, 64, 8) & 2        & 2           \\
% $28^2 \times 64$        & TBS         & (1, 4)    & (8, 128, 8)& 2        & 1      \\
% $14^2 \times 128$       & TBS         & (1, 4)    & (8, 128, 8)& 2        & 2           \\
% $14^2 \times 128$       & TBS         & (1, 4)    & (8, 224, 8)& 4        & 1      \\
% $14^2 \times 224$       & TBS         & (1, 4)    & (8, 368, 8)& 2        & 1      \\
% $7^2 \times 368$        & TBS         & (1, 4)    & (8, 368, 8)& 2        & 2           \\
% $7^2 \times 368$        & 1x1         & -         & 1984       & 1        & 1           \\
% $7^2 \times 1984$       & avgpool     & -         & -          & 1        & 1           \\
% $1984$                  & fc          & -         & 1000       & 1        & -      \\ \bottomrule
% \end{tabular*}

\begin{tabular*}{0.48\textwidth}{l @{\extracolsep{\fill}} lllll}
\toprule
Max. Input              & b           & e         & f          & n        & s \\ \midrule
$256^2 \times 3$          & 3x3         & 1         & 16         & 1        & 2 \\

$128^2 \times 16$ 	 & TBS 	 & 1 	 & (12, 16, 4) 	 & 1 	 & 1 \\
$128^2 \times 16$ 	 & TBS 	 & (0.75, 3.25, 0.5) 	 & (16, 28, 4) 	 & 1 	 & 2 \\
$64^2 \times 28$ 	 & TBS 	 & (0.75, 3.25, 0.5) 	 & (16, 28, 4) 	 & 2 	 & 1 \\
$64^2 \times 28$ 	 & TBS 	 & (0.75, 3.25, 0.5) 	 & (16, 40, 8) 	 & 1 	 & 2 \\
$32^2 \times 40$ 	 & TBS 	 & (0.75, 3.25, 0.5) 	 & (16, 40, 8) 	 & 2 	 & 1 \\
$32^2 \times 40$ 	 & TBS 	 & (0.75, 3.75, 0.5) 	 & (48, 96, 8) 	 & 1 	 & 2 \\
$16^2 \times 96$ 	 & TBS 	 & (0.75, 3.75, 0.5) 	 & (48, 96, 8) 	 & 2 	 & 1 \\
$16^2 \times 96$ 	 & TBS 	 & (0.75, 4.5, 0.75) 	 & (72, 128, 8) 	 & 4 	 & 1 \\
$16^2 \times 128$ 	 & TBS 	 & (0.75, 4.5, 0.75) 	 & (112, 216, 8) 	 & 1 	 & 2 \\
$8^2 \times 216$ 	 & TBS 	 & (0.75, 4.5, 0.75) 	 & (112, 216, 8) 	 & 3 	 & 1 \\
$8^2 \times 216$        & 1x1         & -         & 1984       & 1        & 1           \\
$8^2 \times 1984$       & avgpl     & -         & -          & 1        & 1           \\
$1984$                  & fc          & -         & 1000       & 1        & -      \\ \bottomrule
\end{tabular*}

\label{tab:macro-space}
\end{table}

\begin{table}[]
\small
\centering
\caption{\textbf{Micro-architecture} search space for block design: non-linearities, kernel sizes, and Squeeze-and-Excite~\cite{hu2018squeeze}.}
\begin{tabular*}{0.48\textwidth}{l @{\extracolsep{\fill}} lll}
\toprule
block type          & kernel     & squeeze-and-excite & non-linearity           \\ \midrule
ir\_k3              & 3     & N  & relu        \\
ir\_k5              & 5     & N  & relu        \\
ir\_k3\_hs          & 3     & N  & hswish      \\
ir\_k5\_hs          & 5     & N  & hswish      \\
ir\_k3\_se          & 3     & Y  & relu    \\
ir\_k5\_se          & 5     & Y  & relu    \\
ir\_k3\_se\_hs      & 3     & Y  & hswish      \\
ir\_k5\_se\_hs      & 5     & Y  & hswish      \\
skip                & -     & -  & -   \\ \bottomrule
\end{tabular*}
\label{tab:micro-space}
\end{table}

% skipped block figure for now
Table~\ref{tab:micro-space} describes the micro-architecture search space: the block structure is inspired by~\cite{mobilenetv2, mobilenetv3} and sequentially consists of a $1\times1$ point-wise convolution, a $3\times3$ or $5\times5$ depth-wise convolution, and another $1\times1$ point-wise convolution. Table~\ref{tab:macro-space} describes the macro-architecture. The search space contains more than $10^{35}$ candidate architectures, which is $10^{14}\times$ larger than DNAS's~\cite{fbnet}.

\subsection{Memory Cost}

\begin{figure}
    \centering
    \includegraphics[width=0.23\textwidth]{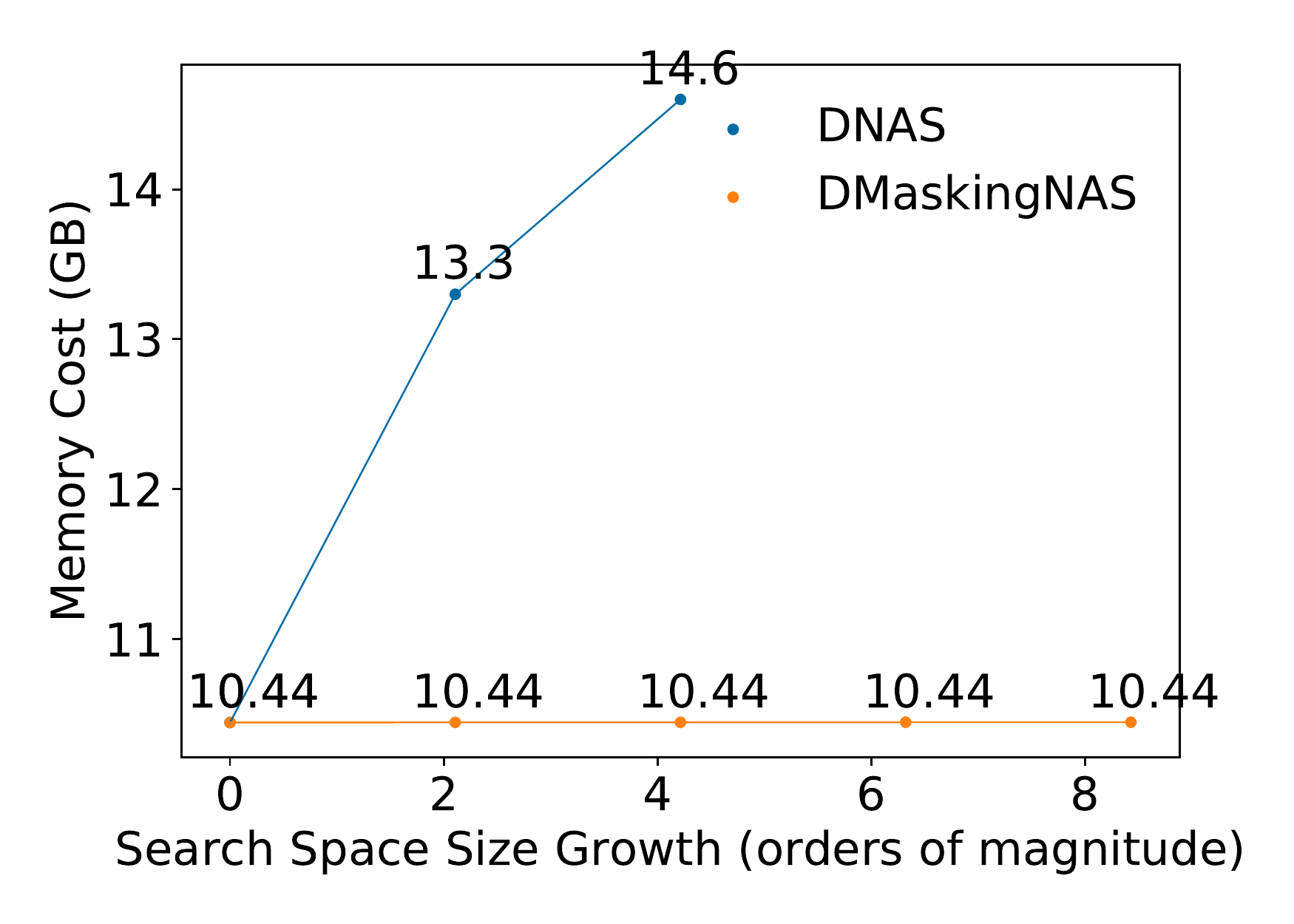}
    \includegraphics[width=0.23\textwidth]{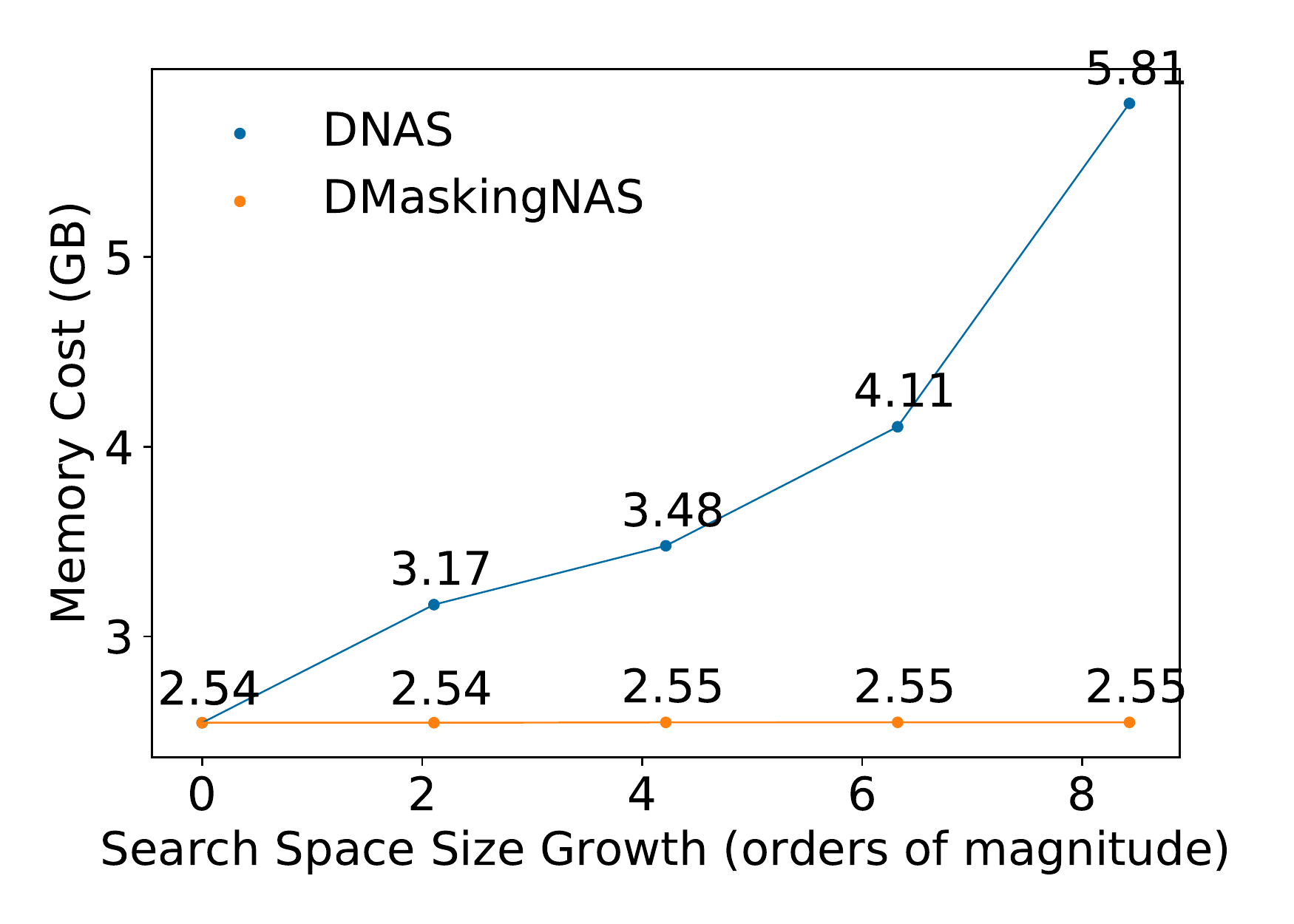}
    \caption{\textbf{Memory Cost of DNAS vs. DMaskingNAS} (Left) Conventional DNAS does not fit into memory with just 8 options per block in channel search. On the other hand, DMaskingNAS's memory cost remains roughly constant, even with 32 channel options per block. (Right) We reduce the number of block options in the search space to fit conventional DNAS into memory. The memory cost growth, as the search space increases, is significantly steeper than that of DMaskingNAS; in fact, DMaskingNAS's memory cost is nearly constant.}
    \label{fig:memory-cost}
\end{figure}

Our memory optimizations yield a $\sim$1MB increase in memory cost for every 2 orders of magnitude the channel search space grows by; for context, this 1 MB increase is just ~0.1\% of the total memory cost during training. This is due to our feature map reuse as described in Sec.~\ref{sec:channel-search}. We compare memory costs for DNAS and DMaskingNAS as the number of channel options increases (Fig.~\ref{fig:memory-cost}, left). With only 8 channel options for each convolution, DNAS fails to fit in memory during training, exceeding the 16GB memory supported by a Tesla V100 GPU. On the other hand, DMaskingNAS supports 32-option channel search, for a $32^{22} \sim 10^{33}$ in search space size (given our 22-layer search space), at nearly constant memory cost. Here, $k$-option channel search means that for each convolution with $c$ channels, we search over $\{c/k, 2c/k, ..., c\}$ channels. To compare larger numbers of channel options, we reduce the number of blocks options in the search space (Fig.~\ref{fig:memory-cost}, right). To compute memory cost, we average the maximum memory allocated during each training step, across 10 epochs.

\subsection{Search for ImageNet Models}

\begin{figure}
    \centering
    \includegraphics[width=0.46\textwidth]{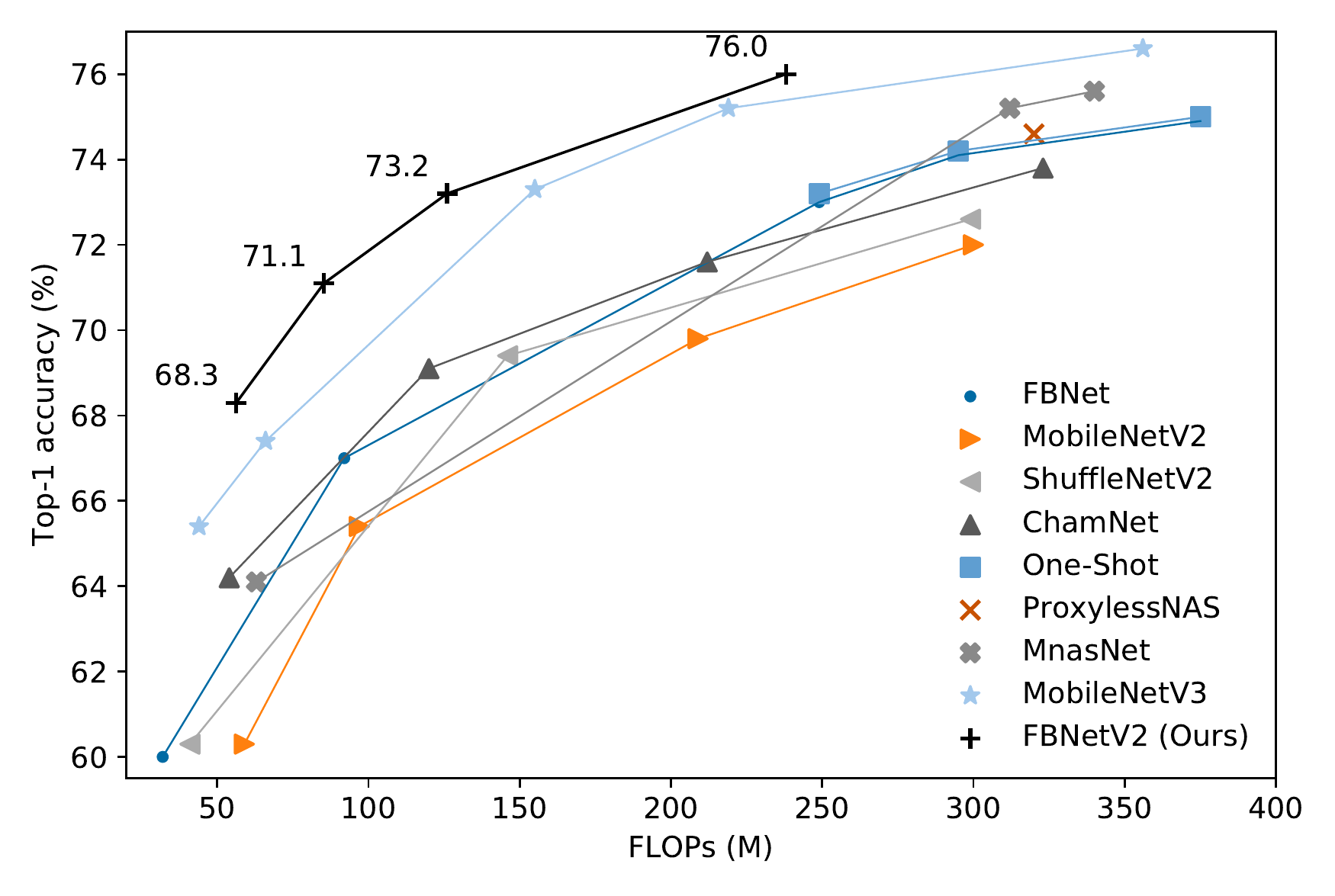}
    \caption{\textbf{ImageNet Accuracy vs. Model FLOPs.} We refer to these FLOP-efficient FBNetV2s as FBNetV2-F\{1, 2, 3, 4\} from left to right.} 
    \label{fig:results-flop-efficient}
\end{figure}

\textbf{FLOP-efficient models}: We first use DMaskingNAS to find compact models (Fig.~\ref{fig:architectures}) for low computational budgets, with models ranging from 50 MFLOPs to 300 MFLOPs in Fig.~\ref{fig:results-flop-efficient}. The searched FBNetV2s outperform all existing networks.

% give some numbers
% not enough space? can revisit

\begin{table*}[t]
\small
\centering
\begin{tabular*}{\textwidth}{l @{\extracolsep{\fill}} lllll}
\toprule
Model & \multicolumn{3}{c}{Search} & FLOPs & Top-1 \\ 
 \cmidrule{2-4}
 & Method & Space & Cost (GPU hours) & & Acc (\%) \\
\midrule
MobileNetV2-$0.35\times$~\cite{mobilenetv2} & manual & - & - & 59M & 60.3 \\
ShuffleNetV2-$0.5\times$~\cite{shufflenetv2} & manual  & - & - & 41M  & 60.3 \\
MnasNet-$0.35\times$~\cite{mnasnet} & RL  & stage-wise & 91K$^*$ & 63M & 64.1    \\
ChamNet-E~\cite{chamnet}& EA & stage-wise  & 28K$\dag$ & 54M & 64.2 \\
FBNet-$0.35\times$~\cite{fbnet}  & gradient & layer-wise  & 0.2K  & 72M & 65.3 \\ 
MobileNetV3-Small~\cite{mobilenetv3} & RL/NetAdapt & stage-wise & $>$91K$\ddag$  & 66M & 67.4 \\
\textbf{FBNetV2-F1 (ours)} & \textbf{gradient} & \textbf{layer-wise}  & \textbf{0.2K} & \textbf{56M}    & \textbf{68.3}   \\ \hline
%1.0-MobileNetV2 \cite{mobilenetv2} & manual & - & - & 300M & 72.0 \\
%1.5-ShuffleNetV2 \cite{shufflenetv2} & manual  & - & - & 299M    & 72.6 \\
%CondenseNet (G=C=8) \cite{condensenet}& manual  & -  & - & 274M    & 71.0 \\
%MnasNet-65 \cite{mnasnet} & RL  & stage-wise & 91K$^*$ / x & 270M & 73.0    \\ 
%DARTS \cite{darts}  & gradient & cell & 288 / 1.33x & 595M & \textbf{73.1}   \\ 
%FBNetV2-Small (ours) & gradient & layer-wise  & 216 / 1.0x & \textbf{249M}    & 73.0   \\ \hline
% CondenseNet (G=C=8) \cite{condensenet}& manual  & -  & -  & 274M    & 71.0 \\
MobileNetV2-$1.0\times$ \cite{mobilenetv2} & manual & - & - & 300M & 72.0 \\
ShuffleNetV2-$1.5\times$ \cite{shufflenetv2} & manual  & - & - & 299M    & 72.6 \\
DARTS  \cite{darts}  & gradient & cell & 0.3K & 595M & 73.1   \\ 
\textbf{FBNetV2-F3 (ours)} & \textbf{gradient} & \textbf{layer-wise}  & \textbf{0.2K} & \textbf{126M} & \textbf{73.2}   \\
\hline
ChamNet-B~\cite{chamnet}& EA & stage-wise  & 28K$\dag$ & 323M & 73.8 \\
FBNet-B~\cite{fbnet} & gradient & layer-wise & 0.2K & 295M & 74.1 \\
One-Shot NAS~\cite{one-shot} & EA & layer-wise & 0.3K & 295M & 74.2 \\
% they mentioned 12+1 GPU days
ProxylessNAS~\cite{proxylessnas} & gradient/RL  & layer-wise & 0.2K & 320M & 74.6 \\
MobileNetV3-Large~\cite{mobilenetv3} & RL/NetAdapt & stage-wise & $>$91K$\ddag$ & 219M & 75.2 \\
MnasNet-A1 \cite{mnasnet} & RL  & stage-wise & 91K$^*$ & 312M & 75.2    \\ 
\textbf{FBNetV2-F4 (ours)} & \textbf{gradient} & \textbf{layer-wise}  & \textbf{0.2K} & \textbf{238M} & \textbf{76.0}   \\ \hline
ResNet-50~\cite{resnet} & manual & - & - & 4.1B & 76.0 \\
DenseNet-169~\cite{densenet} & manual & - & - & 3.5B & 76.2 \\
EfficientNet-B0~\cite{efficientnet} & RL/scaling & stage-wise & $>$91K$\ddag$ & 390M & 77.3 \\
\textbf{FBNetV2-L1 (ours)} & \textbf{gradient} & \textbf{layer-wise} & \textbf{0.6K} & \textbf{325M} & \textbf{77.2} \\ \bottomrule
\end{tabular*}
\caption{\small{\textbf{ImageNet classification performance}: For baselines, we cite statistics on ImageNet from the original papers. Our results are bolded. $^*$: The search cost is estimated based on the experimental setup in~\cite{mnasnet}.  $\dag$:~\cite{chamnet} discovers 5 models with the cost of training 240 networks. $\ddag$: The cost estimation is a lower bound. ~\cite{mobilenetv3} and~\cite{efficientnet} combines the approach proposed in~\cite{mnasnet} with~\cite{netadapt} and compound scaling.}}
\label{tab:imagenet}

\end{table*}

\begin{figure}
    \centering
    \includegraphics[width=0.46\textwidth]{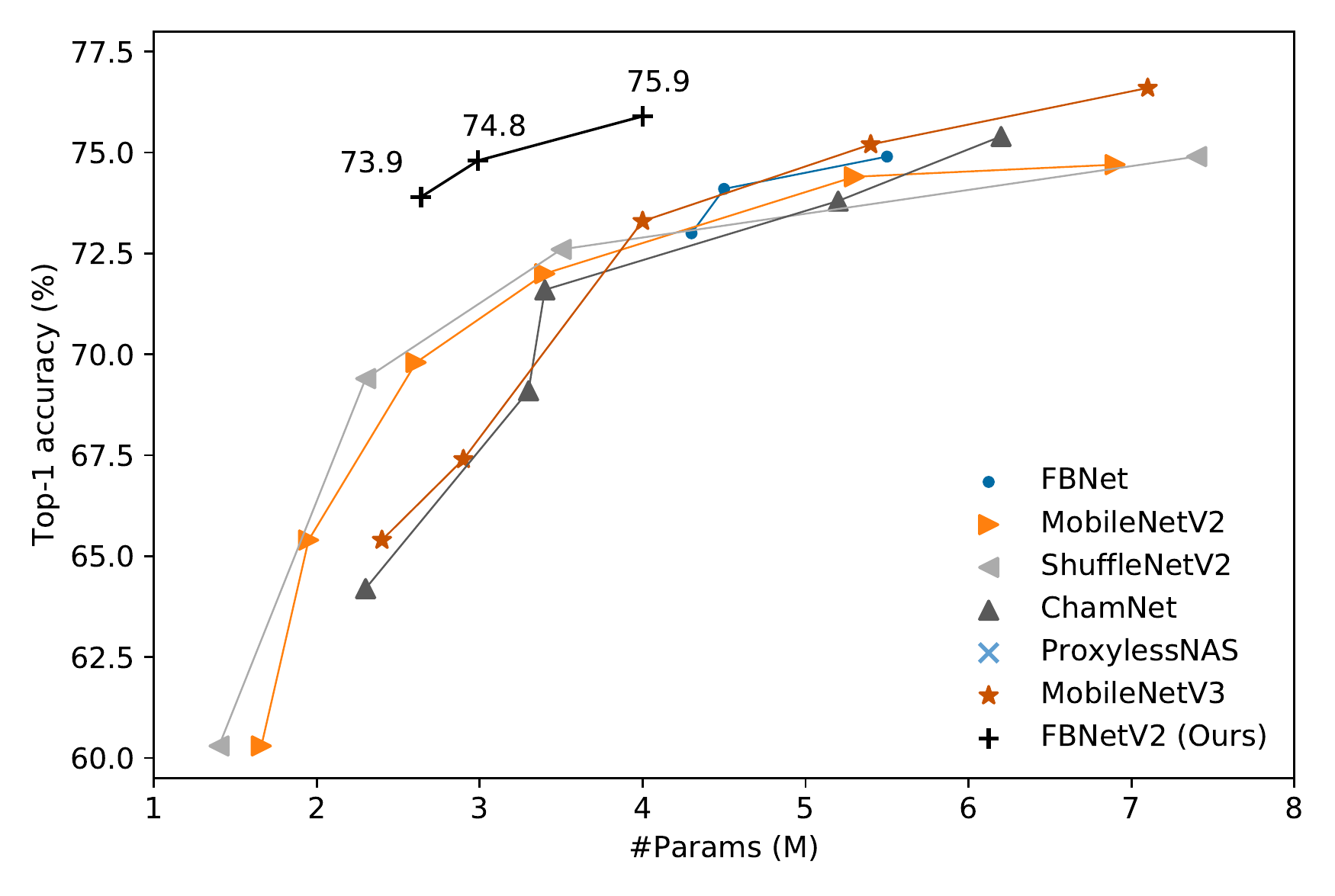}
    \caption{\textbf{ImageNet Accuracy vs. Model Size.} We refer to these as parameter-efficient FBNetV2s as FBNetV2-P\{1, 2, 3\} from left to right.}
    \label{fig:results-param-efficient}
\end{figure}

\textbf{Storage-efficient models}: 
Many real world scenarios face limited on-device storage space. Thus, we next perform searches for models minimizing parameter count, in Fig.~\ref{fig:results-param-efficient}. With similar or smaller model size (4M parameters), FBNetV2 achieves 2.6\% and 2.9\% absolute accuracy gains over MobileNetV3~\cite{mobilenetv3} and FBNet~\cite{fbnet}, respectively.

\begin{figure}
    \centering
    \includegraphics[width=0.46\textwidth]{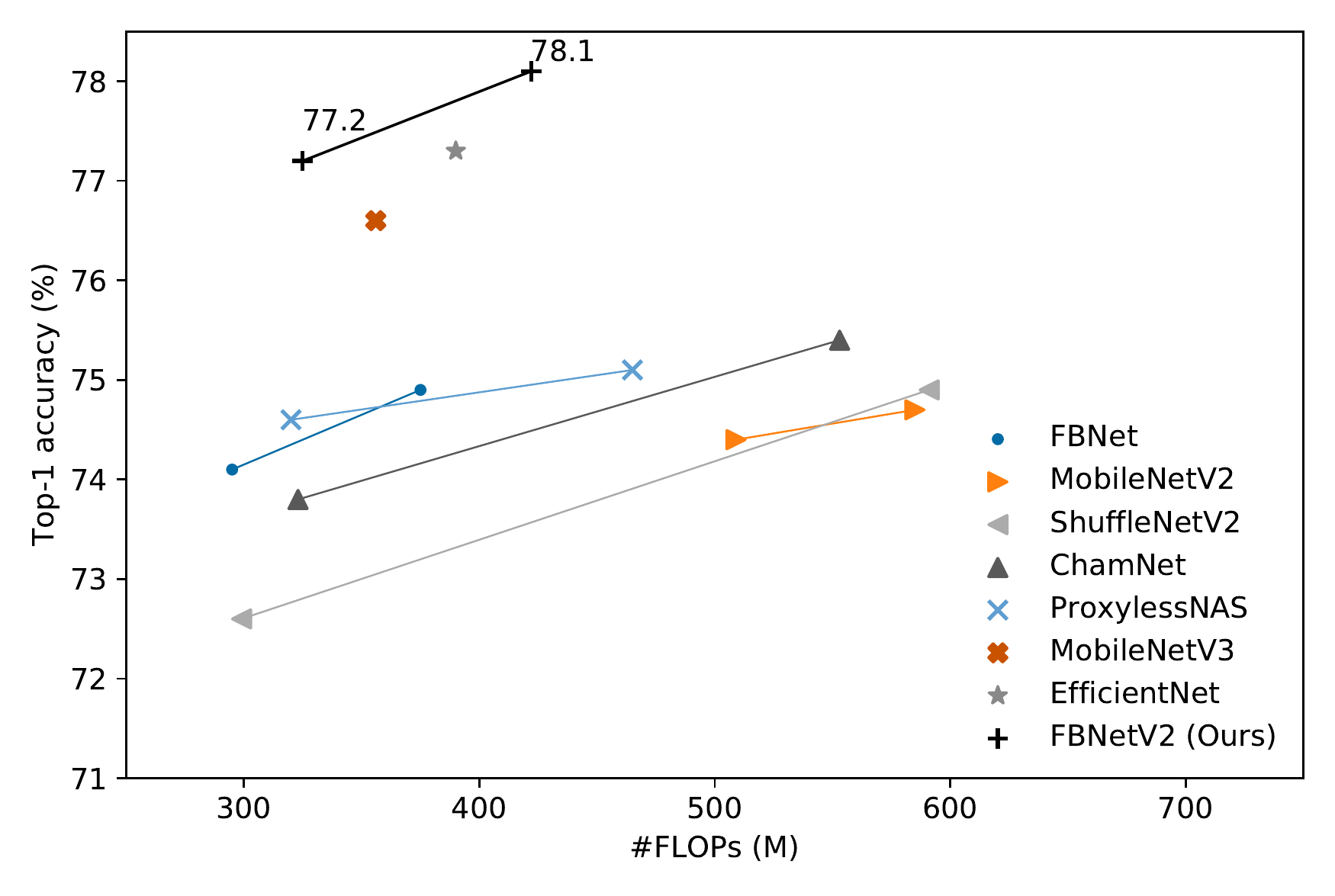}
    \caption{\textbf{ImageNet Accuracy vs. Model FLOPs for Large Models.} We refer to these large FBNetV2s as FBNetV2-L\{1, 2\} from left to right.}
    \label{fig:results-large}
\end{figure}

\textbf{Large models}: 
We finally use DMaskingNAS to explore larger models for high-end devices. We compare FBNetV2-Large with networks of 300+ MFLOPs in Fig.~\ref{fig:results-large}.

% \section{Discussion}

% \subsection{Wedge bottleneck}
% add a discussion about the wedge shaped bottleneck (i.e., \#Channels 192, 184, 176...) ?

\section{Conclusions}
We propose a memory-efficient algorithm, drastically expanding the search space for DNAS by supporting searches over spatial and channel dimensions. These contributions target the main bottleneck for DNAS -- high memory cost that induces constraints on the search space size -- and yield state-of-the-art performance.

\textbf{Acknowledgements} In addition to NSF CISE Expeditions Award CCF-1730628, UC Berkeley research is supported by gifts from Alibaba, Amazon Web Services, Ant Financial, CapitalOne, Ericsson, Facebook, Futurewei, Google, Intel, Microsoft, Nvidia, Scotiabank, Splunk and VMware. This material is based upon work supported by the National Science Foundation Graduate Research Fellowship under Grant No. DGE 1752814.

{\small
\bibliographystyle{ieee_fullname}
\bibliography{bib}

\begin{thebibliography}{10}\itemsep=-1pt

\bibitem{nvda_guide}
Deep learning performance guide.
\newblock
  \url{https://docs.nvidia.com/deeplearning/sdk/dl-performance-guide/index.html}.

\bibitem{proxylessnas}
Han Cai, Ligeng Zhu, and Song Han.
\newblock Proxylessnas: {Direct} neural architecture search on target task and
  hardware.
\newblock {\em arXiv preprint arXiv:1812.00332}, 2018.

\bibitem{chamnet}
Xiaoliang Dai, Peizhao Zhang, Bichen Wu, Hongxu Yin, Fei Sun, Yanghan Wang,
  Marat Dukhan, Yunqing Hu, Yiming Wu, Yangqing Jia, et~al.
\newblock Chamnet: Towards efficient network design through platform-aware
  model adaptation.
\newblock In {\em Proceedings of the IEEE Conference on Computer Vision and
  Pattern Recognition}, pages 11398--11407, 2019.

\bibitem{imagenet}
Jia Deng, Wei Dong, Richard Socher, Li-Jia Li, Kai Li, and Li Fei-Fei.
\newblock {ImageNet}: {A} large-scale hierarchical image database.
\newblock In {\em Proc. IEEE Conf. Computer Vision and Pattern Recognition},
  pages 248--255, 2009.

\bibitem{nassurvey}
Thomas Elsken, Jan~Hendrik Metzen, and Frank Hutter.
\newblock Neural architecture search: {A} survey.
\newblock {\em arXiv preprint arXiv:1808.05377}, 2018.

\bibitem{one-shot}
Zichao Guo, Xiangyu Zhang, Haoyuan Mu, Wen Heng, Zechun Liu, Yichen Wei, and
  Jian Sun.
\newblock Single path one-shot neural architecture search with uniform
  sampling.
\newblock {\em arXiv preprint arXiv:1904.00420}, 2019.

\bibitem{deepcompression}
Song Han, Huizi Mao, and William~J Dally.
\newblock Deep compression: {Compressing} deep neural networks with pruning,
  trained quantization and {Huffman} coding.
\newblock {\em arXiv preprint arXiv:1510.00149}, 2015.

\bibitem{netprune}
Song Han, Jeff Pool, John Tran, and William Dally.
\newblock Learning both weights and connections for efficient neural network.
\newblock In {\em Proc. Advances in Neural Information Processing Systems},
  pages 1135--1143, 2015.

\bibitem{resnet}
Kaiming He, Xiangyu Zhang, Shaoqing Ren, and Jian Sun.
\newblock Deep residual learning for image recognition.
\newblock In {\em Proc. IEEE Conf. Computer Vision and Pattern Recognition},
  pages 770--778, 2016.

\bibitem{he2018sfp}
Yang He, Guoliang Kang, Xuanyi Dong, Yanwei Fu, and Yi Yang.
\newblock Soft filter pruning for accelerating deep convolutional neural
  networks.
\newblock {\em arXiv preprint arXiv:1808.06866}, 2018.

\bibitem{mobilenetv3}
Andrew Howard, Mark Sandler, Grace Chu, Liang-Chieh Chen, Bo Chen, Mingxing
  Tan, Weijun Wang, Yukun Zhu, Ruoming Pang, Vijay Vasudevan, et~al.
\newblock Searching for mobilenetv3.
\newblock {\em arXiv preprint arXiv:1905.02244}, 2019.

\bibitem{mobilenet}
Andrew~G Howard, Menglong Zhu, Bo Chen, Dmitry Kalenichenko, Weijun Wang,
  Tobias Weyand, Marco Andreetto, and Hartwig Adam.
\newblock {MobileNets:} {Efficient} convolutional neural networks for mobile
  vision applications.
\newblock {\em arXiv preprint arXiv:1704.04861}, 2017.

\bibitem{hu2018squeeze}
Jie Hu, Li Shen, and Gang Sun.
\newblock Squeeze-and-excitation networks.
\newblock In {\em Proceedings of the IEEE conference on computer vision and
  pattern recognition}, pages 7132--7141, 2018.

\bibitem{densenet}
Gao Huang, Zhuang Liu, Laurens Van Der~Maaten, and Kilian~Q Weinberger.
\newblock Densely connected convolutional networks.
\newblock In {\em Proceedings of the IEEE conference on computer vision and
  pattern recognition}, pages 4700--4708, 2017.

\bibitem{squeezenet}
Forrest~N Iandola, Song Han, Matthew~W Moskewicz, Khalid Ashraf, William~J
  Dally, and Kurt Keutzer.
\newblock {SqueezeNet}: {AlexNet-level} accuracy with 50x fewer parameters and
  $<$0.5 {MB} model size.
\newblock {\em arXiv preprint arXiv:1602.07360}, 2016.

\bibitem{gumbel}
Eric Jang, Shixiang Gu, and Ben Poole.
\newblock Categorical reparameterization with gumbel-softmax.
\newblock {\em arXiv preprint arXiv:1611.01144}, 2016.

\bibitem{kingma2014adam}
Diederik~P Kingma and Jimmy Ba.
\newblock Adam: A method for stochastic optimization.
\newblock {\em arXiv preprint arXiv:1412.6980}, 2014.

\bibitem{autodeeplab}
Chenxi Liu, Liang-Chieh Chen, Florian Schroff, Hartwig Adam, Wei Hua, Alan~L.
  Yuille, and Li Fei-Fei.
\newblock Auto-deeplab: Hierarchical neural architecture search for semantic
  image segmentation.
\newblock In {\em The IEEE Conference on Computer Vision and Pattern
  Recognition (CVPR)}, June 2019.

\bibitem{progressive}
Chenxi Liu, Barret Zoph, Jonathon Shlens, Wei Hua, Li-Jia Li, Li Fei-Fei, Alan
  Yuille, Jonathan Huang, and Kevin Murphy.
\newblock Progressive neural architecture search.
\newblock {\em arXiv preprint arXiv:1712.00559}, 2017.

\bibitem{darts}
Hanxiao Liu, Karen Simonyan, and Yiming Yang.
\newblock Darts: {Differentiable} architecture search.
\newblock {\em arXiv preprint arXiv:1806.09055}, 2018.

\bibitem{autoslim}
Ning Liu, Xiaolong Ma, Zhiyuan Xu, Yetang Wang, Jian Tang, and Jieping Ye.
\newblock Autoslim: An automatic dnn structured pruning framework for
  ultra-high compression rates, 07 2019.

\bibitem{liu2017slimming}
Zhuang Liu, Jianguo Li, Zhiqiang Shen, Gao Huang, Shoumeng Yan, and Changshui
  Zhang.
\newblock Learning efficient convolutional networks through network slimming.
\newblock In {\em Proceedings of the IEEE International Conference on Computer
  Vision}, pages 2736--2744, 2017.

\bibitem{shufflenetv2}
Ningning Ma, Xiangyu Zhang, Hai-Tao Zheng, and Jian Sun.
\newblock {ShuffleNet V2}: {Practical} guidelines for efficient {CNN}
  architecture design.
\newblock {\em arXiv preprint arXiv:1807.11164}, 2018.

\bibitem{enas}
Hieu Pham, Melody~Y Guan, Barret Zoph, Quoc~V Le, and Jeff Dean.
\newblock Efficient neural architecture search via parameter sharing.
\newblock {\em arXiv preprint arXiv:1802.03268}, 2018.

\bibitem{evolution}
Esteban Real, Sherry Moore, Andrew Selle, Saurabh Saxena, Yutaka~Leon Suematsu,
  Jie Tan, Quoc~V Le, and Alexey Kurakin.
\newblock Large-scale evolution of image classifiers.
\newblock In {\em Proceedings of the 34th International Conference on Machine
  Learning-Volume 70}, pages 2902--2911. JMLR. org, 2017.

\bibitem{mobilenetv2}
Mark Sandler, Andrew Howard, Menglong Zhu, Andrey Zhmoginov, and Liang-Chieh
  Chen.
\newblock Inverted residuals and linear bottlenecks: {Mobile} networks for
  classification, detection and segmentation.
\newblock {\em arXiv preprint arXiv:1801.04381}, 2018.

\bibitem{single_path}
Dimitrios Stamoulis, Ruizhou Ding, Di Wang, Dimitrios Lymberopoulos, Bodhi
  Priyantha, Jie Liu, and Diana Marculescu.
\newblock Single-path nas: Designing hardware-efficient convnets in less than 4
  hours.
\newblock {\em arXiv preprint arXiv:1904.02877}, 2019.

\bibitem{singlepathnas}
Dimitrios Stamoulis, Ruizhou Ding, Di Wang, Dimitrios Lymberopoulos, Bodhi
  Priyantha, Jie Liu, and Diana Marculescu.
\newblock Single-path nas: Device-aware efficient convnet design, 05 2019.

\bibitem{mnasnet}
Mingxing Tan, Bo Chen, Ruoming Pang, Vijay Vasudevan, and Quoc~V Le.
\newblock {MnasNet}: {Platform-aware} neural architecture search for mobile.
\newblock {\em arXiv preprint arXiv:1807.11626}, 2018.

\bibitem{efficientnet}
Mingxing Tan and Quoc~V Le.
\newblock Efficientnet: Rethinking model scaling for convolutional neural
  networks.
\newblock {\em arXiv preprint arXiv:1905.11946}, 2019.

\bibitem{haq}
Kuan Wang, Zhijian Liu, Yujun Lin, Ji Lin, and Song Han.
\newblock Haq: Hardware-aware automated quantization with mixed precision.
\newblock In {\em Proceedings of the IEEE Conference on Computer Vision and
  Pattern Recognition}, pages 8612--8620, 2019.

\bibitem{structured_sparsity}
Wei Wen, Chunpeng Wu, Yandan Wang, Yiran Chen, and Hai Li.
\newblock Learning structured sparsity in deep neural networks.
\newblock In {\em Proc. Advances in Neural Information Processing Systems},
  pages 2074--2082, 2016.

\bibitem{fbnet}
Bichen Wu, Xiaoliang Dai, Peizhao Zhang, Yanghan Wang, Fei Sun, Yiming Wu,
  Yuandong Tian, Peter Vajda, Yangqing Jia, and Kurt Keutzer.
\newblock Fbnet: {Hardware-aware} efficient convnet design via differentiable
  neural architecture search.
\newblock In {\em Proceedings of the IEEE Conference on Computer Vision and
  Pattern Recognition}, pages 10734--10742, 2019.

\bibitem{squeezedet}
Bichen Wu, Forrest Iandola, Peter~H Jin, and Kurt Keutzer.
\newblock Squeezedet: {ified}, small, low power fully convolutional neural
  networks for real-time object detection for autonomous driving.
\newblock In {\em Proceedings of the IEEE Conference on Computer Vision and
  Pattern Recognition Workshops}, pages 129--137, 2017.

\bibitem{snas}
Sirui Xie, Hehui Zheng, Chunxiao Liu, and Liang Lin.
\newblock {SNAS}: {Stochastic} neural architecture search.
\newblock {\em arXiv preprint arXiv:1812.09926}, 2018.

\bibitem{energy_aware}
Tien-Ju Yang, Yu-Hsin Chen, and Vivienne Sze.
\newblock Designing energy-efficient convolutional neural networks using
  energy-aware pruning.
\newblock {\em arXiv preprint arXiv:1611.05128}, 2016.

\bibitem{netadapt}
Tien-Ju Yang, Andrew Howard, Bo Chen, Xiao Zhang, Alec Go, Mark Sandler,
  Vivienne Sze, and Hartwig Adam.
\newblock {NetAdapt}: {Platform-aware} neural network adaptation for mobile
  applications.
\newblock In {\em Proc. European Conf. Computer Vision}, volume~41, page~46,
  2018.

\bibitem{slimmable}
Jiahui Yu, Linjie Yang, Ning Xu, Jianchao Yang, and Thomas Huang.
\newblock Slimmable neural networks.
\newblock In {\em International Conference on Learning Representations}, 2019.

\bibitem{NASRL}
Barret Zoph and Quoc~V Le.
\newblock Neural architecture search with reinforcement learning.
\newblock {\em arXiv preprint arXiv:1611.01578}, 2016.

\bibitem{nasnet}
Barret Zoph, Vijay Vasudevan, Jonathon Shlens, and Quoc Le.
\newblock Learning transferable architectures for scalable image recognition.
\newblock pages 8697--8710, 06 2018.

\end{thebibliography}
}

\end{document}

%%%%%%%%% TITLE
\title{FBNetV2 Supplementary Materials}
\author{}

\maketitle
%\thispagestyle{empty}

%%%%%%%%% BODY TEXT
\section{FBNetV2 on ImageNet}

We include numeric results for all three categories of FBNetV2s, optimized for various resource constraints: FLOP-efficient FBNetV2-F and large FBNetV2-L in Table~\ref{tab:flop}, parameter-efficient FBNetV2-P in Table~\ref{tab:param}. See the main manuscript for comparison with previously state-of-the-art results.

\begin{table}
\small
\centering
\begin{tabular*}{0.48\textwidth}{l @{\extracolsep{\fill}} lll}
\toprule

Model & Input & Flops & Top-1 (\%) \\ 
\midrule
FBNetV2-F1 & 128 & 56M    & 68.3   \\
FBNetV2-F2 & 160 & 85M    & 71.1   \\
FBNetV2-F3 & 192 & 126M    & 73.2   \\
FBNetV2-F4 & 224 & 238M    & 76.0   \\
\hline
FBNetV2-L1 & 224 & 325M & 77.2 \\ 
FBNetV2-L2 & 256 & 422M & 78.1 \\
\bottomrule
\end{tabular*}
\caption{\small{\textbf{ImageNet FLOP-efficient classification}: These are the FBNetV2 models yielded by DMaskingNAS optimizing for FLOP count and accuracy.}}
\label{tab:flop}

\end{table}

\begin{table}
\small
\centering
\begin{tabular*}{0.48\textwidth}{l @{\extracolsep{\fill}} lll}
\toprule

Model & Input & Params & Top-1 (\%) \\ 
\midrule
FBNetV2-P1 & 288 & 2.64M    & 73.9   \\
FBNetV2-P2 & 288 & 2.99M    & 74.8   \\
FBNetV2-P3 & 288 & 4.00M    & 75.9   \\
\bottomrule
\end{tabular*}
\caption{\small{\textbf{ImageNet parameter-efficient classification}: These are the FBNetV2 models yielded by DMaskingNAS optimizing for parameter count and accuracy.}}
\label{tab:param}

\end{table}

\section{Macro-architecture Search Spaces}

We list the DMaskingNAS macro-architecture search spaces for all three categories of FBNetV2s, optimized for various resource constraints: FLOP-efficient FBNetV2-F in Table~\ref{tab:search-space-f}, parameter-efficient FBNetV2-P in Table~\ref{tab:search-space-p}, and large FBNetV2-L in Table~\ref{tab:search-space-l}. Note that in all classes of models, the micro-architecture search space over blocks remains the same.

\begin{table}[h]
\footnotesize
\centering
\caption{Macro-architecture for our largest search space for \textbf{FBNetV2-L}, describing block type $b$, block expansion rate $e$, number of filters $f$, number of blocks $n$. ``TBS'' means layer type needs to be searched. Tuples of three values additionally represent steps between options (low, high, steps). The maximum input resolution for FBNetV2-L is 256.
}

\begin{tabular*}{0.48\textwidth}{l @{\extracolsep{\fill}} lllll}
\toprule
Max. Input              & b           & e         & f          & n        & s \\ \midrule
$256^2 \times 3$          & 3x3         & 1         & 16         & 1        & 2 \\

$128^2 \times 16$ 	 & TBS 	 & 1 	 & (12, 16, 4) 	 & 1 	 & 1 \\
$128^2 \times 16$ 	 & TBS 	 & (0.75, 3.25, 0.5) 	 & (16, 28, 4) 	 & 1 	 & 2 \\
$64^2 \times 28$ 	 & TBS 	 & (0.75, 3.25, 0.5) 	 & (16, 28, 4) 	 & 2 	 & 1 \\
$64^2 \times 28$ 	 & TBS 	 & (0.75, 3.25, 0.5) 	 & (16, 40, 8) 	 & 1 	 & 2 \\
$32^2 \times 40$ 	 & TBS 	 & (0.75, 3.25, 0.5) 	 & (16, 40, 8) 	 & 2 	 & 1 \\
$32^2 \times 40$ 	 & TBS 	 & (0.75, 3.75, 0.5) 	 & (48, 96, 8) 	 & 1 	 & 2 \\
$16^2 \times 96$ 	 & TBS 	 & (0.75, 3.75, 0.5) 	 & (48, 96, 8) 	 & 2 	 & 1 \\
$16^2 \times 96$ 	 & TBS 	 & (0.75, 4.5, 0.75) 	 & (72, 128, 8) 	 & 4 	 & 1 \\
$16^2 \times 128$ 	 & TBS 	 & (0.75, 4.5, 0.75) 	 & (112, 216, 8) 	 & 1 	 & 2 \\
$8^2 \times 216$ 	 & TBS 	 & (0.75, 4.5, 0.75) 	 & (112, 216, 8) 	 & 3 	 & 1 \\
$8^2 \times 216$        & 1x1         & -         & 1984       & 1        & 1           \\
$8^2 \times 1984$       & avgpl     & -         & -          & 1        & 1           \\
$1984$                  & fc          & -         & 1000       & 1        & -      \\ \bottomrule
\end{tabular*}
\label{tab:search-space-l}
\end{table}

\begin{table}[h]
\footnotesize
\centering
\caption{Macro-architecture for our FLOP-efficient search space for \textbf{FBNetV2-F}. The maximum input resolution for FBNetV2-F is 224. See Table \ref{tab:search-space-l} for column names.
}
\begin{tabular*}{0.48\textwidth}{l @{\extracolsep{\fill}} lllll}
\toprule
Max. Input              & b           & e         & f          & n        & s \\ \midrule
$224^2 \times 3$          & 3x3         & 1         & 16         & 1        & 2 \\

$112^2 \times 16$ 	 & TBS 	 & 1 	 & (12, 16, 4) 	 & 1 	 & 1 \\
$112^2 \times 16$ 	 & TBS 	 & (0.75, 4.5, 0.75) 	 & (16, 24, 4) 	 & 1 	 & 2 \\
$56^2 \times 24$ 	 & TBS 	 & (0.75, 4.5, 0.75) 	 & (16, 24, 4) 	 & 2 	 & 1 \\
$56^2 \times 24$ 	 & TBS 	 & (0.75, 4.5, 0.75) 	 & (16, 40, 8) 	 & 1 	 & 2 \\
$28^2 \times 40$ 	 & TBS 	 & (0.75, 4.5, 0.75) 	 & (16, 40, 8) 	 & 2 	 & 1 \\
$28^2 \times 40$ 	 & TBS 	 & (0.75, 4.5, 0.75) 	 & (48, 80, 8) 	 & 1 	 & 2 \\
$14^2 \times 80$ 	 & TBS 	 & (0.75, 4.5, 0.75) 	 & (48, 80, 8) 	 & 2 	 & 1 \\
$14^2 \times 80$ 	 & TBS 	 & (0.75, 4.5, 0.75) 	 & (72, 112, 8) 	 & 3 	 & 1 \\
$14^2 \times 112$ 	 & TBS 	 & (0.75, 4.5, 0.75) 	 & (112, 184, 8) 	 & 1 	 & 2 \\
$7^2 \times 184$ 	 & TBS 	 & (0.75, 4.5, 0.75) 	 & (112, 184, 8) 	 & 3 	 & 1\\
$7^2 \times 184$        & 1x1         & -         & 1984       & 1        & 1           \\
$7^2 \times 1984$       & avgpl     & -         & -          & 1        & 1           \\
$1984$                  & fc          & -         & 1000       & 1        & -      \\ \bottomrule
\end{tabular*}
\label{tab:search-space-f}
\end{table}

\begin{table}[h]
\footnotesize
\centering
\caption{Macro-architecture for our parameter-efficient search space for \textbf{FBNetV2-P}. The maximum input resolution for FBNetV2-P is 288. See Table \ref{tab:search-space-l} for column names.
}
\begin{tabular*}{0.48\textwidth}{l @{\extracolsep{\fill}} lllll}
\toprule
Max. Input              & b           & e         & f          & n        & s \\ \midrule
$288^2 \times 3$          & 3x3         & 1         & 32         & 1        & 2 \\
$144^2 \times 16$ 	 & TBS 	 & 1 	 & (16, 28, 4) 	 & 1 	 & 1 \\
$144^2 \times 28$ 	 & TBS 	 & (0.75, 4.5, 0.75) 	 & (16, 40, 4) 	 & 1 	 & 2 \\
$72^2 \times 40$ 	 & TBS 	 & (0.75, 4.5, 0.75) 	 & (16, 40, 4) 	 & 2 	 & 1 \\
$72^2 \times 40$ 	 & TBS 	 & (0.75, 4.5, 0.75) 	 & (16, 48, 8) 	 & 1 	 & 2 \\
$36^2 \times 48$ 	 & TBS 	 & (0.75, 4.5, 0.75) 	 & (16, 48, 8) 	 & 2 	 & 1 \\
$36^2 \times 48$ 	 & TBS 	 & (0.75, 4.5, 0.75) 	 & (48, 96, 8) 	 & 1 	 & 2 \\
$18^2 \times 96$ 	 & TBS 	 & (0.75, 4.5, 0.75) 	 & (48, 96, 8) 	 & 2 	 & 1 \\
$18^2 \times 96$ 	 & TBS 	 & (0.75, 4.5, 0.75) 	 & (72, 128, 8) 	 & 4 	 & 1 \\
$18^2 \times 128$ 	 & TBS 	 & (0.75, 4.5, 0.75) 	 & (112, 216, 8) 	 & 1 	 & 2 \\
$9^2 \times 216$ 	 & TBS 	 & (0.75, 4.5, 0.75) 	 & (112, 216, 8) 	 & 3 	 & 1\\
$9^2 \times 216$ 	 & TBS 	 & (0.75, 4.5, 0.75) 	 & (112, 216, 8) 	 & 1 	 & 1 \\
$9^2 \times 216$        & 1x1         & -         & 1280       & 1        & 1           \\
$9^2 \times 1280$       & avgpl     & -         & -          & 1        & 1           \\
$1280$                  & fc          & -         & 1000       & 1        & -      \\ \bottomrule
\end{tabular*}
\label{tab:search-space-p}
\end{table}

% \hyperlink{https://github.com/facebookresearch/mobile-vision}{\color{blue}{https://github.com/facebookresearch/mobile-vision}}.

% \includegraphics[]{}